\def\year{2020}\relax
\documentclass[letterpaper]{article} 
\usepackage{aaai20}  
\usepackage{times}  
\usepackage{helvet} 
\usepackage{courier}  
\usepackage[hyphens]{url}  
\usepackage{graphicx} 
\usepackage{graphicx}  
\usepackage{amsmath,amssymb,amsfonts}
\usepackage{textcomp}
\usepackage{multirow}
\usepackage{xcolor}
\usepackage{subfigure}
\usepackage{algorithm}
\usepackage{algpseudocode}
\usepackage{epsfig}

\title{Hyperbolic Graph Attention Network}
\author{Yiding Zhang\textsuperscript{1}, Xiao Wang\textsuperscript{1}, Xunqiang Jiang\textsuperscript{1}, Chuan Shi\textsuperscript{1}, Yanfang Ye\textsuperscript{2} \\
	\textsuperscript{1}Beijing University of Posts and Telecommunications, Beijing, China \\
	\textsuperscript{2}West Virginia University, WV, USA \\
	\{zyd, xiaowang, skd621, shichuan\}@bupt.edu.cn, yanfang.ye@mail.wvu.edu\\
}

\begin{document}

\maketitle

\begin{abstract}
Graph neural network (GNN) has shown superior performance in dealing with graphs, which has attracted considerable research attention recently.
However, most of the existing GNN models are primarily designed for graphs in Euclidean spaces.
Recent research has proven that the graph data exhibits non-Euclidean latent anatomy.
Unfortunately, there was rarely study of GNN in non-Euclidean settings so far.
To bridge this gap, in this paper, we study the GNN with attention mechanism in hyperbolic spaces at the first attempt.
The research of hyperbolic GNN has some unique challenges: since the hyperbolic spaces are not vector spaces, the vector operations (e.g., vector addition, subtraction, and scalar multiplication) cannot be carried.
To tackle this problem, we employ the gyrovector spaces, which provide an elegant algebraic formalism for hyperbolic geometry, to transform the features in a graph; and then we propose the hyperbolic proximity based attention mechanism to aggregate the features.
Moreover,
as mathematical operations in hyperbolic spaces could be more complicated than those in Euclidean spaces,
we further devise a novel acceleration strategy using logarithmic and exponential mappings to improve the efficiency of our proposed model.
The comprehensive experimental results on four real-world datasets demonstrate the performance of our proposed hyperbolic graph attention network model,
by comparisons with other state-of-the-art baseline methods.
\end{abstract}

\section{Introduction}\label{sec_intro}
The real-world data usually come together with the graph structure, such as social networks, citation networks, biology networks.
Graph neural network (GNN) \cite{gori2005new,scarselli2009graph}, as a powerful deep representation learning method for such graph data, has shown superior performance on network analysis and aroused considerable research interest.
There have been many studies using a neural network to handle the graph data.
For examples, \cite{gori2005new,scarselli2009graph} leveraged deep neural network to learn node representations based on node features and the graph structure;
\cite{defferrard2016convolutional,kipf2016semi,hamilton2017inductive} proposed the graph convolutional networks by generalizing the convolutional operation to graph;
\cite{velivckovic2017graph} designed a novel convolution-style graph neural network by employing the attention mechanism in GNN.
These proposed GNNs have been widely used to solve many real-world application problems, such as recommendation \cite{ying2018graph,song2019session} and disease prediction \cite{parisot2017spectral}.


Essentially, most of the existing GNN models are primarily designed for the graphs in Euclidean spaces.
The main reason is that Euclidean space is the natural generalization of our intuition-friendly and visible three-dimensional space.
However, some researchers have discovered that graph data exhibits a non-Euclidean latent anatomy \cite{wilson2014spherical,bronstein2017geometric}.
In such cases, the Euclidean spaces may not provide the most powerful or meaningful geometry for graph representation learning.
On the other hand, some recent works \cite{krioukov2010hyperbolic,nickel2017poincare} have demonstrated that the hyperbolic spaces could be the latent spaces of graph data, as the hyperbolic space may reflect some properties of graph naturally, e.g., hierarchical and power-law structure \cite{krioukov2010hyperbolic,muscoloni2017machine}.
Inspired by this insight, the study of graph data in hyperbolic spaces has received increasing attention, such as hyperbolic graph embedding \cite{nickel2017poincare,nickel2018learning,de2018representation,wang2019hyperbolic}.%

\begin{figure}
	\centering
	\subfigure[\scriptsize ``Circle Limit 1'', by M.C Escher]{
		\includegraphics[width=0.2\textwidth]{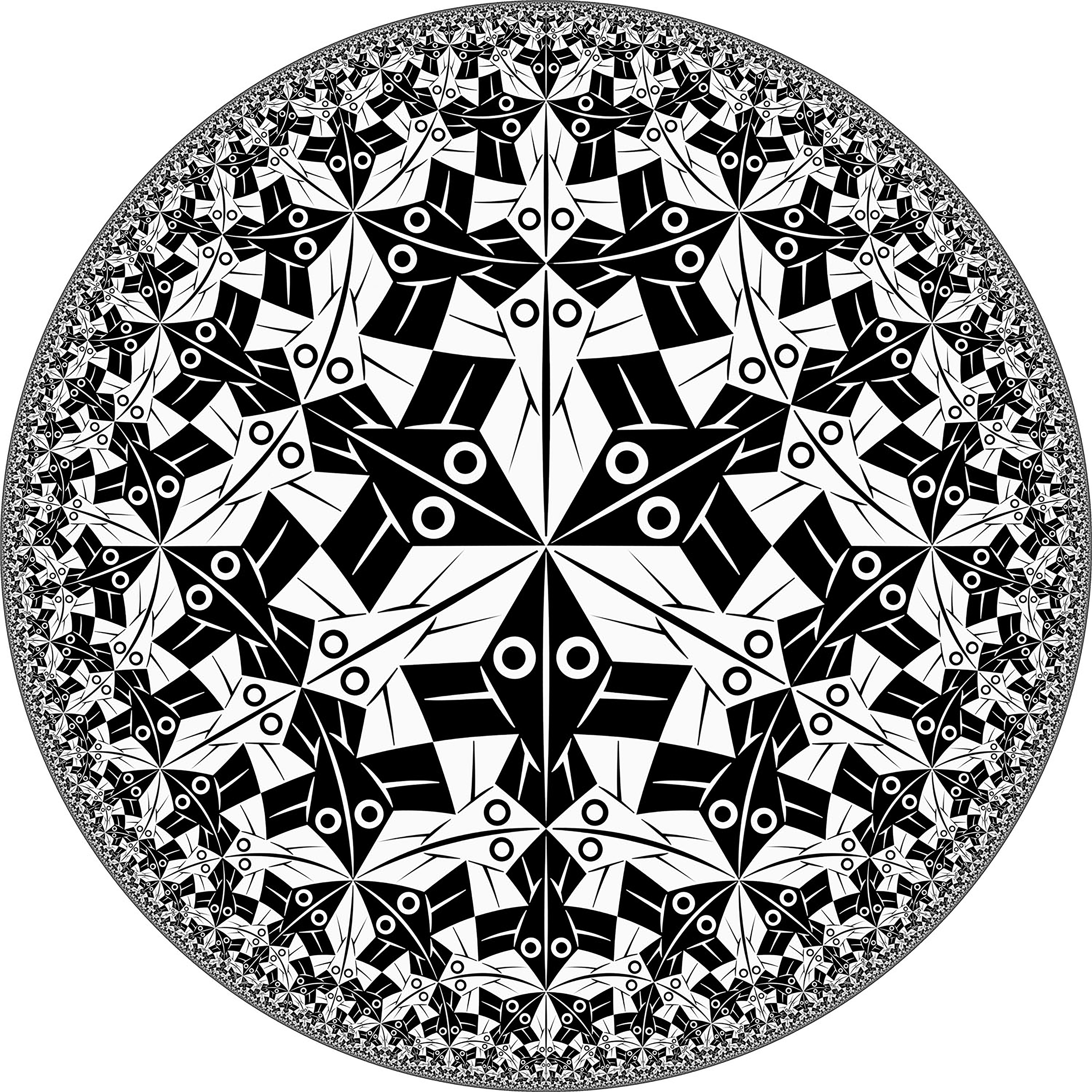}
		\label{fig_poincare_circle_limit1}
	}
	\subfigure[\scriptsize A tree with branching factor 3]{
		\includegraphics[width=0.2\textwidth]{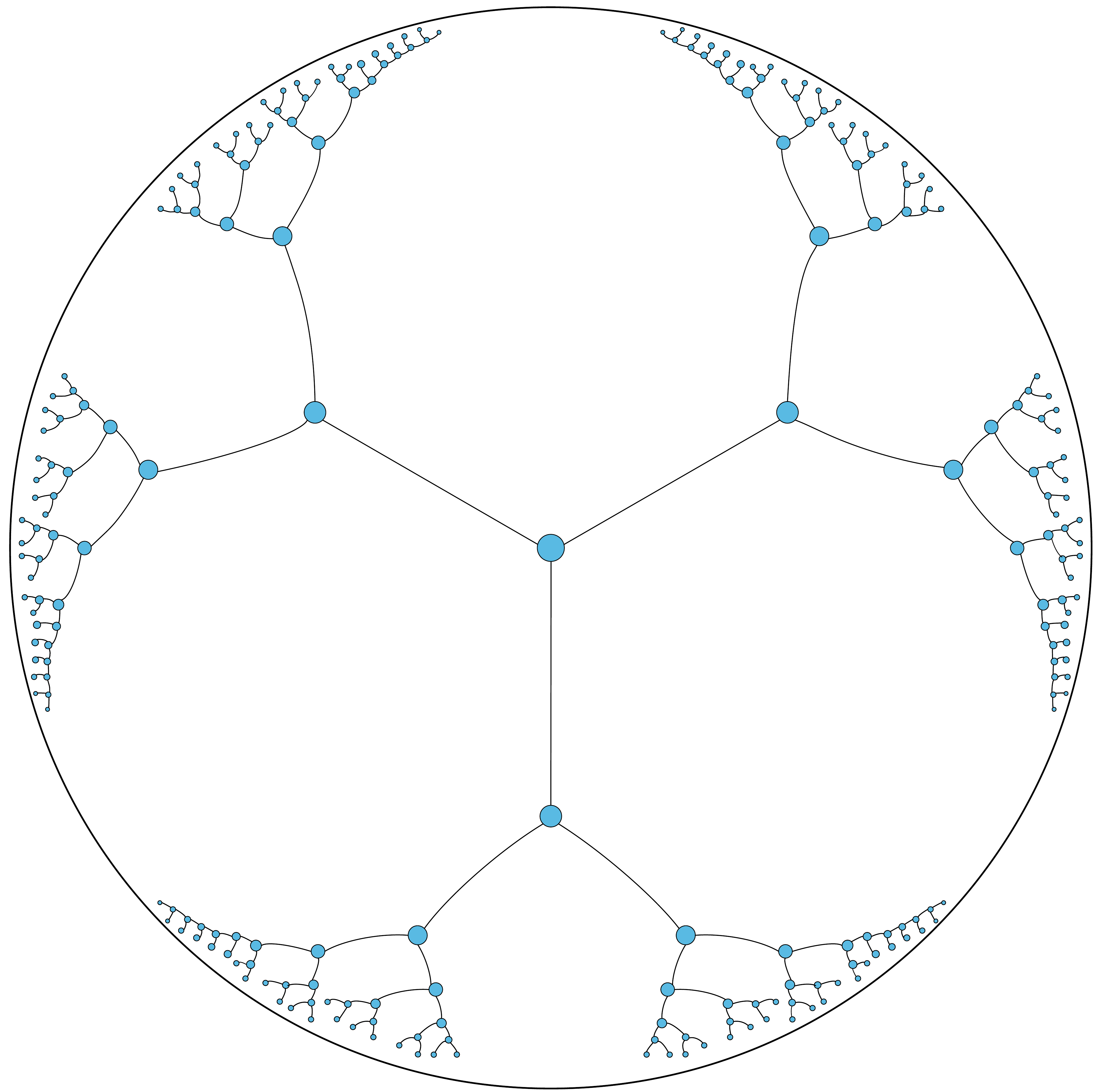}
		\label{fig_poincare_tree}
	}
	\caption{Two examples of hyperbolic spaces (Poincar\'e disk model).}
	\label{fig_poincare}
\end{figure}

One key property of hyperbolic spaces is that they expand faster than Euclidean spaces,
because Euclidean spaces expand polynomially while hyperbolic spaces expand exponentially.
For instance, each tile in Fig. \ref{fig_poincare_circle_limit1} is of equal area in hyperbolic space but diminishes towards zero in Euclidean space towards the boundary.
As the tiles grow exponentially, there is sufficient room to embed these tiles,
so that we have shrunk the tiles in this Euclidean diagram for visualization.
With these properties, hyperbolic spaces can be thought of as ``continue tree''.
As shown in Fig. \ref{fig_poincare_tree}, considering a tree with branching factor $b$,
the number of nodes at level $l$ or no more than $l$ hops from the root are $(b+1)b^{l-1}$ and $[(b+1)b^l - 2]/(b-1)$ respectively.
The number of nodes grows exponentially with their distance to the root of the tree, which is similar to hyperbolic spaces as they expand exponentially.
Therefore, there is a strong correlation between tree-likeness graph and hyperbolic spaces \cite{krioukov2010hyperbolic,nickel2017poincare}.
With this property, hyperbolic spaces have been considered to model complex network recently \cite{krioukov2010hyperbolic,muscoloni2017machine}.
These researches discover that graphs with hierarchical structure and power-law distribution are suitable to be modeled in hyperbolic spaces.
Meanwhile, graph data with these properties exist widely, such as social networks, network community structures, citation networks and biology networks \cite{clauset2009power,krioukov2010hyperbolic},
which motivates us to study the GNN in hyperbolic spaces.

Despite the powerful modeling ability on graph data of hyperbolic spaces,
there are two key challenges in designing the GNN in hyperbolic spaces:
(1) One is that there are many different procedures in GNNs, e.g., the projection step, the attention mechanism, and the propagation step.
However, different from Euclidean spaces, hyperbolic spaces are not vector spaces,
so the vector operations (such as vector addition, and subtraction) cannot be carried in hyperbolic spaces.
\textit{How can we effectively implement those procedures of GNN in hyperbolic spaces in an elegant way?}
(2) Another challenge is that, as the hyperbolic spaces have constant negative curvature,
mathematical operations in hyperbolic spaces could be more complex than those in Euclidean spaces.
Some basic properties of mathematical operations, such as the commutative or associative of ``vector addition''
are not satisfied anymore in hyperbolic spaces.
\textit{How can we assure the learning efficiency in the proposed model?}


To address the above challenges, in this paper, we propose a novel Hyperbolic graph ATtention network (denoted as HAT).
Specifically, we use the framework of gyrovector spaces to build the graph attentional layer in hyperbolic spaces.
Gyrovector spaces are the mathematical concepts proposed by Ungar \cite{ungar2001hyperbolic,ungar2008gyrovector},
which study hyperbolic geometry in an analogy vector spaces way.
In other words,
just like vector spaces form algebraic formalism for Euclidean geometry,
the framework of gyrovector spaces provides an elegant algebraic formalism for hyperbolic geometry.
Therefore, we use the gyrovector operations in hyperbolic spaces to transform the features of the graph and exploit the proximity in hyperbolic spaces to model the attention mechanism.
To improve the learning efficiency,
we further propose a logarithmic mapping and exponential mapping based method to accelerate our model,
in the premise of preserving the character in hyperbolic spaces.
In sum, the major contributions of this work can be summarized as follows:

\begin{itemize}
	\item To the best of our knowledge, we are the first to study graph attention network in hyperbolic spaces, which is potential to learn better representations in graphs.
	
	\item We propose a novel graph attention network in hyperbolic spaces, named HAT. We employ the framework of gyrovector spaces to implement the graph processing in hyperbolic spaces and design an attention mechanism based on hyperbolic proximity.
	
	\item We design a method to accelerate our model while preserving the property in the hyperbolic spaces by using the logarithmic map and exponential map, which assures the efficiency of our proposed HAT model. 
	
	\item We conduct extensive experiments to evaluate the performance of HAT on four datasets. The results show the superiority of HAT in node classification and node clustering tasks compared with the state-of-the-art methods.
\end{itemize}


\section{Related Work}\label{sec_related}

\subsection{Graph Neural Network}
GNN aims to extend the deep neural network to deal with arbitrary graph-structured data \cite{gori2005new,scarselli2009graph}.
Recently, there is a surge of generalizing convolutions to the graph domain.
\cite{defferrard2016convolutional} utilized K-order Chebyshev polynomials to approximate smooth filters in the spectral domain.
\cite{kipf2016semi} leveraged a localized first-order approximation of spectral graph convolutions to learn the node representations.
\cite{velivckovic2017graph} studied the attention mechanism in GNN, which incorporated the attention mechanism into the propagation step.
To sum up, all these GNNs model graphs in Euclidean spaces so far.

\subsection{Representation Learning in Hyperbolic Spaces}
Recently, representation learning in hyperbolic spaces has received increasing attention.
Specifically, 
\cite{nickel2017poincare,nickel2018learning} focused on learning the hierarchical representation of
a graph.
\cite{ganea2018hyperbolic} embedded the directed acyclic graphs into hyperbolic spaces to learn their feature representations.
Besides, some researchers began to study deep learning in hyperbolic spaces.
\cite{ganea2018hnn} generalized deep neural models in hyperbolic spaces, such as recurrent neural networks and gated recurrent unit.
\cite{gulcehre2018hyperbolic} imposed hyperbolic geometry on the activations of the neural network, while the other structures of this network are in Euclidean spaces.

\section{Preliminaries}\label{sec_prelim}

\subsection{Hyperbolic Spaces and Graph Data}
We provide some detail reasons for modeling graphs with hyperbolic geometry.
As mentioned in Introduction, one key property of hyperbolic spaces is that they expand faster than Euclidean spaces.
Specifically, considering a disk in a 2-dimensional hyperbolic space with constant curvature $K = -1$, 
the perimeter and area of the disk of hyperbolic radius $r$ are given as $2\pi\sinh r$ and $2\pi(\cosh r - 1)$, respectively,
and both of them grow as $e^r$ with $r$.
\footnote{Because of $\sinh r=\frac{1}{2}( e^r - e^{-r})$, $\cosh r=\frac{1}{2}( e^r + e^{-r})$}
In a 2-dimensional Euclidean space,
the length of a circle and the area of a disk of Euclidean radius $r$ are given as $2\pi r$ and $\pi r^2$,
growing only linearly and quadratically about $r$.
With this property, some researches discover that hyperbolic spaces may be the inherent spaces for graphs with hierarchal structure and power-law distribution \cite{krioukov2010hyperbolic,muscoloni2017machine}.
Hence, many real graphs with hierarchical structure and power-law distribution are suitable to be modeled in hyperbolic spaces \cite{papadopoulos2012popularity,faqeeh2018characterizing}.
Moreover, some physical researchers have discovered that this kind of structure is a 
\textit{universal phenomenon} 
for real-world graphs \cite{clauset2008hierarchical},
including citation networks, social networks, biology networks \cite{clauset2009power,krioukov2010hyperbolic}.

\subsection{Gyrovector Spaces}
Vector spaces form the algebraic formalism in Euclidean spaces so that we can use vector operations such as vector addition, subtraction and scalar multiplication in Euclidean spaces.
We are familiar with these operations which can be used to design algorithms in Euclidean space.
However, they cannot be carried in hyperbolic spaces.
Fortunately, just like the vector spaces form the algebraic formalism for Euclidean geometry,
the framework of gyrovector spaces provides an algebraic formalism for hyperbolic geometry \cite{ungar2001hyperbolic,ungar2008gyrovector}.
The gyrovector spaces enable the vector operations, such as vector addition and scalar multiplication, to be carried in hyperbolic spaces.
We can use gyrovector operations to design the algorithms in hyperbolic spaces.
Therefore, we briefly introduce the framework of gyrovector spaces here.

In particular, the operations in gyrovector spaces are defined in an open $d$-dimensional ball:
$$
\mathbb{D}_c^d:=\{\mathbf{x}\in \mathbb{R}^d: c\|\mathbf{x}\|^2<1\},
$$
where $c\ge0$ is corresponding to the radius of the ball.
If $c=0$ i.e., $\mathbb{D}_c^d=\mathbb{R}^d$, the ball equals to the Euclidean space;
if $c>0$, $\mathbb{D}_c^d$ is the open ball of radius $\frac{1}{\sqrt{c}}$;
if $c=1$, we recover the usual ball $\mathbb{D}^d$.
The gyrovector operations are performed in this $d$-dimensional ball.

\section{The Proposed Model}\label{sec_model}

\begin{figure}[htb]
	\centering
	\includegraphics[width=0.45\textwidth]{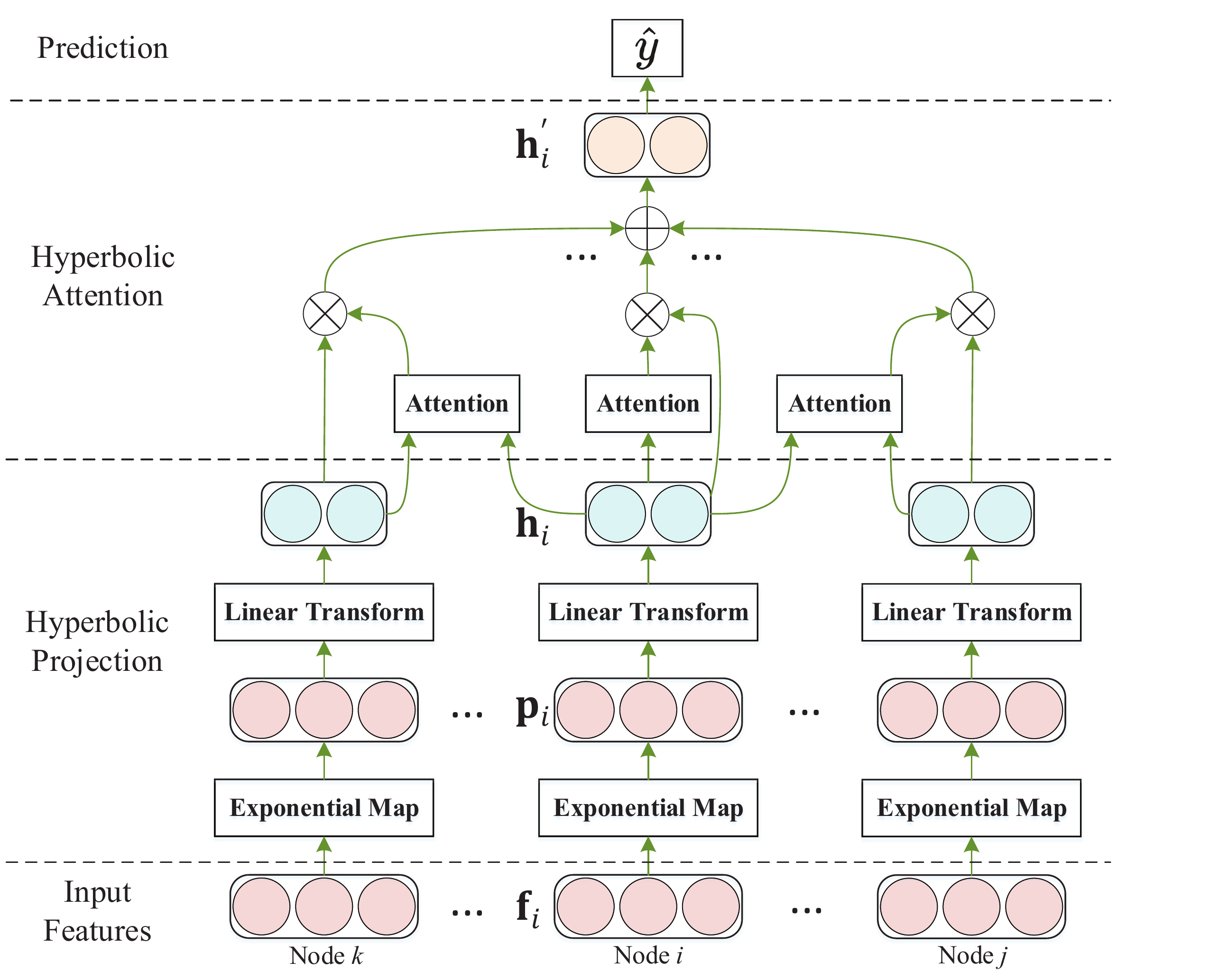}
	\caption{The framework of HAT model.}
	\label{model}
\end{figure}

In this section, we present our hyperbolic graph attention network model, named HAT, whose framework is shown in Fig. \ref{model}.
In general, we should project and transform the input node feature in a hyperbolic space,
and design hyperbolic attention mechanism with the node feature.
Hence, our model can be summarized as two procedures:
(1) \textbf{The hyperbolic feature projection}.
Given the original input node feature, this procedure projects it into
a hyperbolic space through the exponential map and the hyperbolic linear transformation,
so as to obtain the latent representation of the node in hyperbolic space.
(2) \textbf{The hyperbolic attention mechanism}.
This procedure designs an attention mechanism based on the hyperbolic proximity to aggregate the latent representations.
Finally, we feed the aggregated representations to a loss function for the downstream task.
Here we mainly describe a single graph attentional layer, 
as the sole layer is used throughout all of our proposed HAT architectures in our experiments.
Furthermore, we devise an acceleration strategy to speed up the proposed model by using logarithmic and exponential mapping.

\subsection{The  HAT Model}


\subsubsection{The hyperbolic feature projection}

The input of GNN is the node feature, whose norm could be out of the open ball defined in gyrovector spaces.
To make the node feature available in hyperbolic spaces, we use the exponential map to project the feature into the hyperbolic spaces.
Specifically, let $\mathbf{f}_i$ be the feature of node $i$, and then
for $\mathbf{f}_i\in T_{\mathbf{x}}\mathbb{D}_c^d\backslash\{\mathbf{0}\}$,
where $\mathbf{x}$ is a point in hyperbolic spaces and
$T_{\mathbf{x}}\mathbb{D}_c^d$ is the tangent space at point $\mathbf{x}$,
the exponential map $\exp_{\mathbf{x}}^c: T_{\mathbf{x}}\mathbb{D}_c^d\rightarrow\mathbb{D}_c^d$ is given for $\mathbf{x}\neq\mathbf{0}$ by:
\begin{equation}\label{eq_exp_map}
\exp_{x}^c(\mathbf{f}_i)={\mathbf{x}}\oplus_c\Big(\tanh(\sqrt{c}\frac{\lambda_{{\mathbf{x}}}^c \|\mathbf{f}_i\|}{2})\frac{\mathbf{f}_i}{\sqrt{c}\|\mathbf{f}_i\|}\Big),
\end{equation}
when $\mathbf{x}=\mathbf{0}$, the exponential map is defined as:
\begin{equation}\label{eq_exp_map_0}
\exp_{\mathbf{0}}^c(\mathbf{f}_i)=\tanh(\sqrt{c}\|\mathbf{f}_i\|)\frac{\mathbf{f}_i}{\sqrt{c}\|\mathbf{f}_i\|},
\end{equation}
where $\lambda_{\mathbf{x}}^c:=\frac{2}{(1-c\|\mathbf{x}\|^2)}$ is a conformal factor. The operation $\oplus_c$ is the M\"{o}bius addition, and it will be interpreted in Eq. \eqref{eq_addition}.
Here we assume that the feature $\mathbf{f}_i$ lies in the tangent spaces at the point $\mathbf{x}=\mathbf{0}$,
so we can get the new feature $\mathbf{p}_i\in\mathbb{D}^d_c$ in hyperbolic spaces via $\mathbf{p}_i=\exp_0^c(\mathbf{f}_i)$.

We then transform $\mathbf{p}_i$ into a higher-level latent representation $\mathbf{h}_i$ to obtain sufficient representation power.
To achieve this, we use a shared linear transformation parametrized by a weight matrix ${\mathbf{M}\in \mathbb{R}^{d'\times d}}$ (where $d'$ is the dimension of the final representation)
and employ the M\"{o}bius matrix-vector multiplication \cite{ganea2018hnn}.
If $\mathbf{Mp}_i\neq \mathbf{0}$, we have
\begin{equation}
\label{eq_mobiusmvm}
\mathbf{h}_i=\mathbf{M}{\otimes_c} \mathbf{p}_i = \frac{1}{\sqrt{c}}\tanh\bigg(\frac{\|\mathbf{Mp}_i\|}{\|\mathbf{p}_i\|}\tanh^{-1}(\sqrt{c}\|{\mathbf{p}}_i\|)\bigg),
\end{equation}
and if $\mathbf{Mp}_i=\mathbf{0}$, $\mathbf{M}{\otimes_c} \mathbf{p}_i=0$.
Here $\mathbf{h}_i$
can be considered as a latent representation in the hidden layer of HAT.

\subsubsection{The hyperbolic attention mechanism}

We then perform a self-attention  mechanism on the nodes. The attention coefficient $\alpha_{ij}$, which
indicates the importance of node $j$ to node $i$, can be computed as:
\begin{equation}
\alpha_{ij} = f(\mathbf{h}_i,\mathbf{h}_j),
\end{equation}
where $f$ represents the function of computing the attention coefficient.
Here we only compute $\alpha_{ij}$ for nodes $j\in N_i$, where $N_i$ is the neighbors of node $i$ in the graph.
Considering a large attention coefficient $\alpha_{ij}$ for the high similarity of nodes $j$ and $i$,
we define $f$ based on the distance in hyperbolic spaces,
which can measure the similarity between nodes.
Specifically, if the generalized hyperbolic metric tensor conformal to the Euclidean one,
with conformal factor $\lambda_{\mathbf{x}}^c:=\frac{2}{1-c\|\mathbf{x}\|^2}$,
given two node latent representations $\mathbf{h}_i,\mathbf{h}_j\in\mathbb{D}^d_c$, the distance is given by:
\begin{equation}
d_c(\mathbf{h}_i,\mathbf{h}_j)=\frac{2}{\sqrt{c}}\tanh^{-1}(\sqrt{c}\|-\mathbf{h}_i\oplus_c \mathbf{h}_j\|),
\label{eq_distance}
\end{equation}
where the operator $\oplus_c$ is the M\"{o}bius addition in $\mathbb{D}_c^d$ as:
\begin{equation}
\mathbf{h}_i\oplus_c \mathbf{h}_j\!\! :=\!\! \frac{(1 \!+\! 2c\langle \mathbf{h}_i,\!\mathbf{h}_j\rangle\!+\!c\|\mathbf{h}_j\|^2)\mathbf{h}_i \!+\! (1\!-\!c\|\mathbf{h}_i\|^2)\mathbf{h}_j}{1 + 2c\langle \mathbf{h}_i,\!\mathbf{h}_j \rangle + c^2 \|\mathbf{h}_i\|^2 \|\mathbf{h}_j\|^2}.
\label{eq_addition}
\end{equation}
Then, we perform the self-attention coefficient as:
\begin{equation}
\label{eq_coefficient}
\alpha_{ij} = -d_c\big(\mathbf{h}_i,\mathbf{h}_j\big).
\end{equation}
Because the hyperbolic spaces are metric spaces, there are two advantages of using distance in hyperbolic spaces to calculate the self-attention coefficient.
(1) Different from the inner product in Euclidean spaces, the hyperbolic distance meets the triangle inequality, so the self-attention can preserve the transitivity among nodes.
(2) As we can see, the attention coefficient of a given node $i$ with itself
is $\alpha_{ii}=-d_c(\mathbf{h}_i,\mathbf{h}_i)=0$, which is always be the largest over its neighbors.
As the representation should mainly maintain its own characteristics,
this attention coefficient can meet this requirement in mathematics, while some other graph attention networks, e.g., GAT \cite{velivckovic2017graph}, cannot guarantee this.

For all the neighbors of node $i$ (including itself), we should make their attention coefficients easily comparable,
so we normalize them using the softmax function:
\begin{equation}\label{eq_softmax}
w_{ij} = \frac{\exp(\alpha_{ij})}{\sum_{k\in N_i}\exp(\alpha_{ik})}.
\end{equation}


The normalized attention coefficient $w_{ij}$ is used to compute a linear combination of the latent representations of all
the nodes $j\in N_i$. So the final aggregated representation $\mathbf{h}'_i$ for node $i$ is
as follows:
\begin{equation}
\mathbf{h}_i'=\sigma\bigg(\sum_{j\in N_i}\nolimits^{\oplus_c}  w_{ij}\otimes_c\mathbf{h}_j\bigg),
\label{eq_sum}
\end{equation}
where the $\sum^{\oplus_c}$ is the accumulation of M\"{o}bius addition and $\sigma$ is a nonlinearity function defined as ELU. 
The operation $w_{ij}\otimes_c\mathbf{h}_j$ can be realized by
the M\"{o}bius scalar multiplication. For $c>0$, the M\"obius scalar multiplication of $\mathbf{h}_j\in\mathbb{D}^d_c\backslash\{\mathbf{0}\}$ by $w_{ij}\in\mathbb{R}$ is defined as:
\begin{equation}
w_{ij}\otimes_c\mathbf{h}_j:= \frac{1}{\sqrt{c}}\tanh\big(w_{ij}\tanh^{-1}(\sqrt{c}\|\mathbf{h}_j\|)\big)\frac{\mathbf{h}_j}{\|\mathbf{h}_j\|},
\label{eq_mobius_scalar_mul}
\end{equation}
and $w_{ij}\otimes_c\mathbf{0}:=\mathbf{0}$. 

We can apply the final representations to specific tasks and optimize them with different loss functions.
In this paper, we consider the semi-supervised node classification task and use cross-entropy loss function to train our model.


\subsection{Some properties of M\"obius operations} \label{sec_properties}
To help make sense of M\"obius operations, some properties of them will be expounded in this section.
Some M\"obius operations recover the Euclidean operations when $c$ goes to zero.
Specifically, for M\"{o}bius addition and M\"{o}bius scalar multiplication,
we have $\lim_{c\rightarrow 0}\mathbf{h}_i\oplus_c \mathbf{h}_j = \mathbf{h}_i+ \mathbf{h}_j$
and $\lim_{c\rightarrow 0}w_{ij}\otimes_c \mathbf{h}_j = w_{ij}\mathbf{h}_j$, respectively.
Also,
the M\"obius matrix-vector multiplication and M\"{o}bius scalar multiplication satisfy associativity.
They have
$(\mathbf{MM}')\otimes_c\mathbf{p}_i=\mathbf{M}\otimes_c(\mathbf{M}'\otimes_c\mathbf{p}_i)$,
and
$(r_1r_2)\otimes_c\mathbf{h}=r_1\otimes_c(r_2\otimes_c\mathbf{h})$, respectively.
The M\"{o}bius scalar multiplication also satisfies the scalar distributivity
$(r_1+r_2)\otimes_c\mathbf{h}=r_1\otimes_c\mathbf{h}+r_2\otimes_c\mathbf{h}$.
Moreover, in general, the M\"obius addition is neither commutative nor associative, which results in the inefficient problem of Eq. \eqref{eq_sum}.
This problem will be interpreted in the following.




\subsection{Acceleration of HAT} \label{sec_acceleration}
In our proposed model HAT,
the calculation of Eq. \eqref{eq_sum} is very time-consuming, which
seriously affects the efficiency of HAT.
As mentioned before, the M\"{o}bius addition
in Eq. \eqref{eq_sum} is neither commutative nor associative,
meaning that we have to calculate the results by order.
Specifically, we denote $w_{ij}\otimes_c\mathbf{h}_j$ as $\mathbf{v}_{ij}$, so the accumulation term in Eq. \eqref{eq_sum} can be rewritten as follows:
\begin{equation}
\label{cx}
\begin{aligned}
\sum\nolimits_{j\in N_i}^{\oplus_c} \mathbf{v}_{ij} & = \mathbf{v}_{i1} \oplus_c \mathbf{v}_{i2} \oplus_c \mathbf{v}_{i3} \oplus_c \cdots \\
& = \Big( \big((\mathbf{v}_{i1} \oplus_c \mathbf{v}_{i2}) \oplus_c \mathbf{v}_{i3}\big) \oplus_c \cdots   \Big).
\end{aligned}
\end{equation}
As we can see, the calculation of Eq. \eqref{cx} has to be in a serial manner.
It is well known that there are always some hubs which have many edges in a large graph, so the calculation becomes very impractical.

Actually, some operations in gyrovector spaces can be derived with logarithmic map and exponential map.
Taking the M\"{o}bius scalar multiplication as an example, it first uses the logarithmic map to
project the representation into a tangent space, and then multiply the projected representation by a scalar in the tangent space, and finally project it back on the manifold with the exponential map \cite{ganea2018hnn}.
The logarithmic map and the exponential map can move the representation between the two manifolds in a correct manner.
Specifically, for two points $\mathbf{v}_i\in\mathbb{D}_c^d$ and $\mathbf{v}_j\in\mathbb{D}_c^d\backslash\{\mathbf{0}\}$, the logarithmic map $\log_{{v}_i}^c:\mathbb{D}_c^n\rightarrow T_{{v}_i}\mathbb{D}_c^n$ is given for ${\mathbf{v}_j}\neq {\mathbf{v}_i}$ by:
\begin{equation}
\log_{v_i}^c\!({\mathbf{v}_j}) = \frac{2}{\sqrt{c}\lambda_{{v}_i}^c}\!\tanh^{\!-1}\!(\sqrt{c}\|-{\mathbf{v}_i}\oplus_c {\mathbf{v}_j}\|)\frac{-{\mathbf{v}_i}\!\oplus_c\!{\mathbf{v}_j}}{\|-{\mathbf{v}_i}\!\oplus_c\! {\mathbf{v}_j}\|},
\end{equation}
when $\mathbf{v}_i=\mathbf{0}$, we have:
\begin{equation}
\log_{0}^c\!({\mathbf{v}_j}) = \tanh^{-1}\!(\sqrt{c}\|{\mathbf{v}_j}\|)\frac{{\mathbf{v}_j}}{\|{\mathbf{v}_j}\|}.
\end{equation}

The logarithmic map enables us to get the representations $\log_{0}^c\!({\mathbf{v}_j})$
in a tangent space. As the tangent spaces are vector spaces, we can combine the representations, just as we do it in the Euclidean spaces, i.e.,
$\sum_{j\in N_i}\nolimits  \log_\mathbf{0}^c\big(w_{ij}\otimes_c\mathbf{h}_j\big)$.
After the linear combination, we use the exponential map to project the representations back to the hyperbolic spaces,
giving rise to the final representations as:
\begin{equation}
\mathbf{h_i}'=\sigma\bigg(\exp_{\mathbf{0}}^c\Big(\sum_{j\in N_i}\nolimits  \log_\mathbf{0}^c\big(w_{ij}\otimes_c\mathbf{h}_j\big)\Big)\bigg).
\label{eq_sum2}
\end{equation}
Different from the Eq. \eqref{eq_sum}, the accumulation operation in the Eq. \eqref{eq_sum2} is commutative and associative, so it can be computed in a parallel way.
Thus, our model becomes more efficient.

\subsection{Complexity Analysis}
The time complexity of HAT is $O(|V|\cdot d\cdot d'+|E|\cdot d')$, where $d$ and $d'$ are the dimension of input and output features, respectively.
$|V|$ and $|E|$ are the numbers of nodes and edges in the graph, respectively.
The complexity is on par with other GNN methods, such as GAT \cite{velivckovic2017graph} and GCN \cite{kipf2016semi}.

More importantly, our model can also be parallelized.
For example,
with the proposed acceleration strategy, the computation of the aggregated representation 
(i.e., Eq. \eqref{eq_sum2}) can be parallelized across all nodes.
The operations of the self-attention (i.e., Eq. \eqref{eq_coefficient}) can be parallelized across all edges.
Specifically, taking the Cora graph \cite{sen2008collective} as an example,
conducted on a GPU (NVIDIA GTX 1080 Ti),
HAT only costs about 84 seconds to converge with acceleration strategy,
while cannot converge within 12 hours without acceleration strategy.

\section{Experiments}\label{sec_experiment}

\begin{table}
	\caption{Summary of the datasets.}
	\centering
	\label{table_dataset}
	\resizebox{0.46\textwidth}{!}{
		\begin{tabular}{ccccc}
			\hline
			Dataset    			& Cora   & Citeseer & Pubmed & Amazon Photo\\
			\hline
			\# Nodes    		& 2708   & 3327     & 19717  & 7650		\\
			\# Edges    		& 5429   & 4732	    & 44338  & 143663	\\
			\# Features  		& 1433   & 3703     & 500    & 745		\\
			\# Classes  		& 7      & 6        & 3 	 & 8		\\
			\hline
		\end{tabular}
	}
\end{table}

\subsection{Experiments Setup}

\subsubsection{Datasets}
We employ four widely used real-world graphs for evaluations, including Cora, Citeseer, Pubmed \cite{sen2008collective} and Amazon Photo \cite{shchur2018pitfalls}. Their detailed descriptions are summarized in Table \ref{table_dataset}.
In Cora, Citeseer and Pubmed, node represents document and edge represents the citation relation.
In Amazon Photo, node represents product and edge indicates that two goods are frequently bought together.
All the nodes in these datasets correspond to a label and a bag-of-words representation.
For fair comparison, we follow the setting of former literature \cite{yang2016revisiting,kipf2016semi,velivckovic2017graph}:
for each dataset, we use only 20 nodes per class for training, 500 nodes for validation, 1000 nodes for test,
and the training algorithm could access all nodes' features.


\subsubsection{Baselines}
We compare our method with the following state-of-the-art methods:
(1) graph embedding methods, including some Euclidean graph embedding methods, i.e., DeepWalk \cite{perozzi2014deepwalk}, Node2vec \cite{grover2016node2vec}, LINE \cite{tang2015line}, 
and a hyperbolic graph embedding method, i.e., Poincar\'{e}Emb \cite{nickel2017poincare};
(2) some semi-supervised graph neural networks, i.e., GCN \cite{kipf2016semi} and GAT \cite{velivckovic2017graph}.

\begin{table*}[htbp]
	\caption{Quantitative results on the node classification task. The best results are marked by bold numbers.}
	\centering
	\label{table_classification}
	\begin{center}
		\resizebox{0.98 \textwidth}{!}{
			\begin{tabular}{ccc ccc cccc}
				\hline
				Dataset 				 	& Dimension & DeepWalk &Node2vec& LINE(1st) & LINE(2nd)	& Poincar\'{e}Emb	& \quad GCN\quad    & \qquad GAT\qquad		&\qquad HAT\qquad  \\
				\hline
				\multirow{4}{*}{Cora}  		& 2 		& 0.359 	&0.386	&0.255		&0.180		&0.491				&\quad0.452\quad	 &\qquad0.550\qquad	&\qquad\textbf{0.608}\qquad 	\\
				\multirow{4}{*}{}  				& 4 		& 0.566		&0.593	&0.314		&0.324		&0.536				&\quad0.714\quad	 &\qquad0.751\qquad	&\qquad\textbf{0.787}\qquad 	\\
				\multirow{4}{*}{} 	 			& 8 		& 0.605		&0.635	&0.473		&0.335		&0.574				&\quad0.806\quad	 &\qquad0.798\qquad	&\qquad\textbf{0.828}\qquad  	\\
				\multirow{4}{*}{}  				& 16 		& 0.617		&0.645	&0.485		&0.381		&0.642				&\quad0.815\quad	 &\qquad0.819\qquad	&\qquad\textbf{0.831}\qquad 	\\
				\hline
				\multirow{4}{*}{Citeseer}  		& 2 		& 0.257		&0.316	&0.193		&0.180		&0.287				&\quad0.357\quad	&\qquad0.512\qquad	&\qquad\textbf{0.546}\qquad 	\\
				\multirow{4}{*}{}  				& 4 		& 0.401		&0.427	&0.226		&0.243		&0.310				&\quad0.556\quad	&\qquad0.656\qquad	&\qquad\textbf{0.681}\qquad 	\\
				\multirow{4}{*}{} 	 			& 8 		& 0.427		&0.451	&0.261		&0.245		&0.365				&\quad0.679\quad	&\qquad0.697\qquad	&\qquad\textbf{0.712}\qquad 	\\
				\multirow{4}{*}{}  				& 16 		& 0.459		&0.471	&0.307		&0.269		&0.399				&\quad0.704\quad	&\qquad0.704\qquad	&\qquad\textbf{0.719}\qquad 	\\
				\hline
				\multirow{4}{*}{Pubmed} & 2 		& 0.535		&0.565	&0.342		&0.379		&0.614				&\quad0.632\quad	&\qquad0.743\qquad	&\qquad\textbf{0.761}\qquad 	\\
				\multirow{8}{*}{}  				& 4 		& 0.645		&0.669	&0.504		&0.380		&0.629				&\quad0.708\quad	&\qquad0.761\qquad	&\qquad\textbf{0.767}\qquad 	\\
				\multirow{4}{*}{} 	 			& 8 		& 0.672		&0.692	&0.522		&0.423		&0.659				&\quad\textbf{0.786}\quad	&\qquad0.766\qquad	&\qquad{0.781}\qquad 	\\
				\multirow{4}{*}{}  				& 16 		& 0.681		&0.697	&0.529		&0.479		&0.678				&\quad\textbf{0.791}\quad	&\qquad0.770\qquad	&\qquad{0.782}\qquad 	\\
				\hline
				\multirow{4}{*}{Amazon Photo} & 2 		& 0.580		&0.612	&0.240		&0.239		&0.615				&\quad0.319\quad	&\qquad0.309\qquad	&\qquad\textbf{0.629}\qquad 	\\
				\multirow{4}{*}{}  				& 4 		& 0.756		&0.768	&0.321		&0.613		&0.769				&\quad0.559\quad	&\qquad0.686\qquad	&\qquad\textbf{0.782}\qquad 	\\
				\multirow{4}{*}{} 	 			& 8 		& 0.790		&0.803	&0.529		&0.617		&0.777				&\quad0.786\quad	&\qquad0.784\qquad	&\qquad\textbf{0.843}\qquad 	\\
				\multirow{4}{*}{} 				& 16 		& 0.798		&0.808	&0.624		&0.630		&0.788				&\quad0.819\quad	&\qquad0.835\qquad	&\qquad\textbf{0.858}\qquad 	\\
				\hline		
		\end{tabular}}
	\end{center}
\end{table*}

\subsubsection{Parameter Settings}
For all the methods, we carry the experiments in the embedding dimension of 2, 4, 8, 16 (i.e., the number of hidden units in GNN).
For DeepWalk and Node2vec, we set window size as 5, walk length as 80, walks per node as 40. For Poincar\'eEmb, LINE(1st) and LINE(2nd), we set the number of negative samples as \{5, 10\}.
For GAT, because of the limited of dimension,
we carry the experiments of single head attention.
For HAT, we set $c = 1$.
We tune the parameters for all methods via validation data.
Moreover, HAT without acceleration strategy is very time-consuming,
so we did not carry experiments for that case.

\subsection{Experimental Results}
\subsubsection{Node Classification}
Node classification is a basic task widely used to evaluate the effectiveness of representations.
For GCN, GAT, and HAT, they are the semi-supervised models which can be directly used to classify the nodes.
For DeepWalk and Node2vec, we employ KNN classifier with $k=5$ to perform the node classification.
Because the KNN classifier cannot be directly applied to hyperbolic spaces,
for Poincar\'eEmb, we project the representations in the tangent space at $\mathbf{0}$ via $\log_{\mathbf{0}}^c$, and then feed the representations into the classifier.
We report the average accuracy of 10 runs with random weight initialization.

The results are shown in Table \ref{table_classification}.
It is obvious that HAT achieves the best performance in most cases, and its superiority is more significant for the low dimension setting.
Moreover,
we can find that the GNN based methods (i.e., GCN, GAT, and HAT) perform better than other baselines (i.e., DeepWalk, Node2vec, LINE(1st, 2nd), and Poincar\'eEmb) in most cases,
because of combining the graph structure and node features in their models.
Furthermore,
compared to GNN methods in Euclidean spaces (i.e., GCN, GAT), HAT performs better in most cases, especially in low dimension, suggesting the superiority of modeling graph in hyperbolic spaces.
The superiority of hyperbolic spaces is further validated in the comparison of LINE(1st) and Poincar\'eEmb. 
Although both of them preserve the first-order proximity in graphs, the hyperbolic graph embedding method Poincar\'eEmb always perform better than LINE(1st).

\begin{table*}[htbp]
	\caption{Quantitative results on the node clustering task. The best results are marked by bold numbers.}
	\centering
	\label{table_clustering}
	\begin{center}
		\resizebox{0.98\textwidth}{!}{
			\begin{tabular}{ccc ccc cccc c}
				\hline
				Dataset 							 	& Dimension & DeepWalk &Node2vec	& LINE(1st) & LINE(2nd)	& Poincar\'{e}Emb	& \quad GCN\quad    & \qquad GAT\qquad   	&\qquad HAT\qquad\\
				\hline
				\multirow{4}{*}{Cora} 		& 2 		& 0.264		&0.281		&0.075	&0.074		&0.245		&\quad0.341\quad		&\qquad\textbf{0.404}\qquad		&\qquad{0.382}\qquad	 \\
				\multirow{4}{*}{}  				& 4 		& 0.274		&0.292		&0.239	&0.111		&0.329		&\quad0.428\quad		&\qquad0.504\qquad		&\qquad\textbf{0.519}\qquad	 \\
				\multirow{4}{*}{}  				& 8 		& 0.358		&0.365		&0.277	&0.105		&0.395		&\quad0.501\quad		&\qquad0.572\qquad		&\qquad\textbf{0.582}\qquad  \\
				\multirow{4}{*}{}  				& 16 		& 0.404		&0.415		&0.292	&0.119		&0.441		&\quad0.524\quad		&\qquad\textbf{0.584}\qquad		&\qquad{0.581}\qquad   \\
				\hline
				\multirow{4}{*}{Citeseer} 	& 2 		& 0.090		&0.164		&0.048	&0.012		&0.121		&\quad0.248\quad		&\qquad0.315\qquad		&\qquad\textbf{0.321}\qquad  \\
				\multirow{4}{*}{}  				& 4 		& 0.121		&0.169		&0.104	&0.036		&0.160		&\quad0.344\quad		&\qquad0.391\qquad		&\qquad\textbf{0.399}\qquad	\\
				\multirow{4}{*}{}  				& 8 		& 0.156		&0.177		&0.083	&0.043		&0.194		&\quad0.401\quad		&\qquad0.417\qquad		&\qquad\textbf{0.427}\qquad   \\
				\multirow{4}{*}{} 				& 16 		& 0.179		&0.209		&0.092	&0.057		&0.264		&\quad0.426\quad		&\qquad0.430\qquad		&\qquad\textbf{0.439}\qquad \\
				\hline
				\multirow{4}{*}{Pubmed}  		& 2 		& 0.153		&0.195		&0.076	&0.043		&0.206		&\quad0.230\quad		&\qquad0.334\qquad		&\qquad\textbf{0.345}\qquad  \\
				\multirow{4}{*}{}  				& 4 		& 0.162		&0.214		&0.083	&0.036		&0.221		&\quad0.254\quad		&\qquad0.340\qquad		&\qquad\textbf{0.358}\qquad  \\
				\multirow{4}{*}{} 				& 8 		& 0.196		&0.224		&0.102	&0.055		&0.257		&\quad0.242\quad		&\qquad0.343\qquad		&\qquad\textbf{0.386}\qquad  \\
				\multirow{4}{*}{}  				& 16 		& 0.231		&0.286		&0.115	&0.077		&0.284		&\quad0.262\quad		&\qquad0.352\qquad		&\qquad\textbf{0.393}\qquad \\
				\hline		
				\multirow{4}{*}{Amazon Photo} & 2 		& 0.478		&0.489		&0.145	&0.399		&0.499		&\quad0.187\quad		&\qquad0.464\qquad		&\qquad\textbf{0.505}\qquad  \\
				\multirow{4}{*}{}  				& 4 		& 0.578		&0.584		&0.242	&0.370		&0.527		&\quad0.207\quad		&\qquad0.595\qquad		&\qquad\textbf{0.647}\qquad  \\
				\multirow{4}{*}{}  				& 8 		& 0.643		&0.663		&0.363	&0.413		&0.591		&\quad0.240\quad		&\qquad0.636\qquad		&\qquad\textbf{0.672}\qquad  \\
				\multirow{4}{*}{}  				& 16 		& 0.681		&0.710		&0.388	&0.416		&0.626		&\quad0.254\quad		&\qquad0.659\qquad		&\qquad\textbf{0.719}\qquad \\
				\hline		
		\end{tabular}}
	\end{center}
\end{table*}

\begin{figure}[t]
	\centering
	\subfigure[Neighbors of P1728]{
		\includegraphics[width=0.3\textwidth]{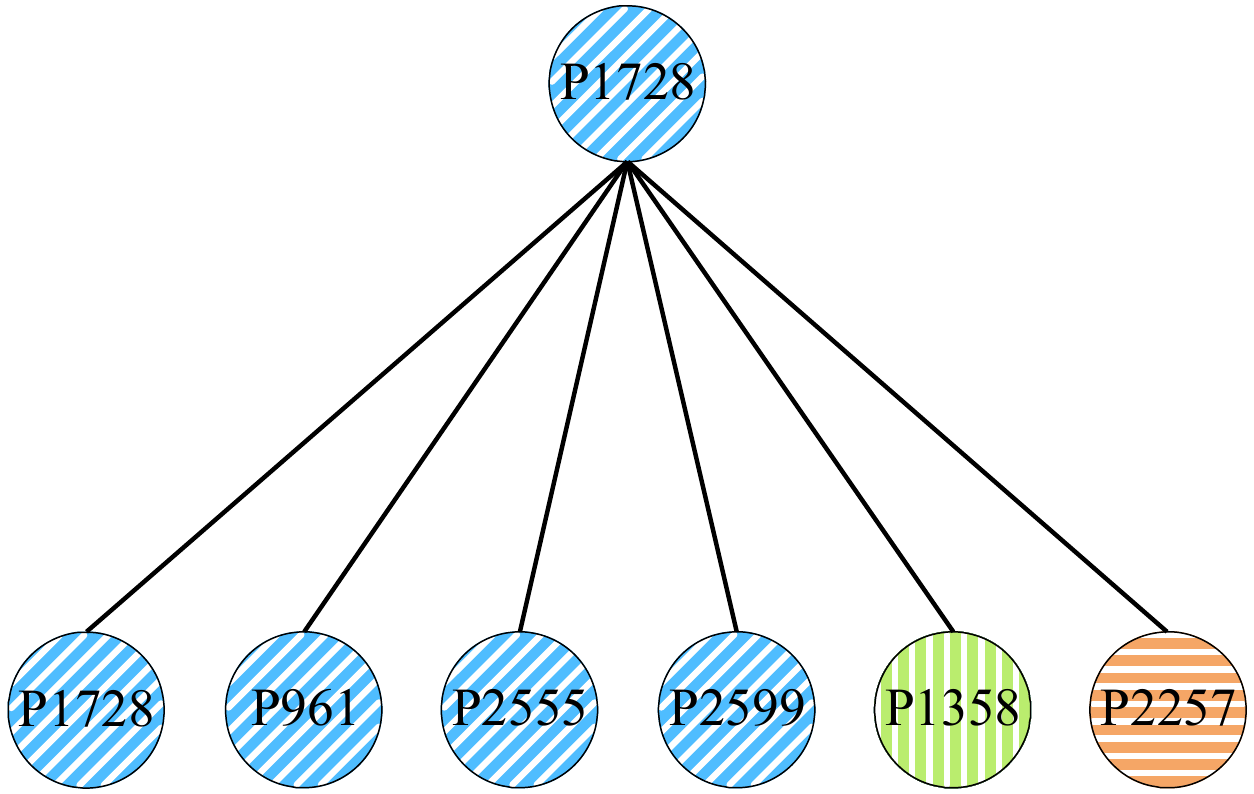}
		\label{fig_attn_1}
	}
	\subfigure[Attention values of P1728's neighbors]{
		\includegraphics[width=0.3\textwidth]{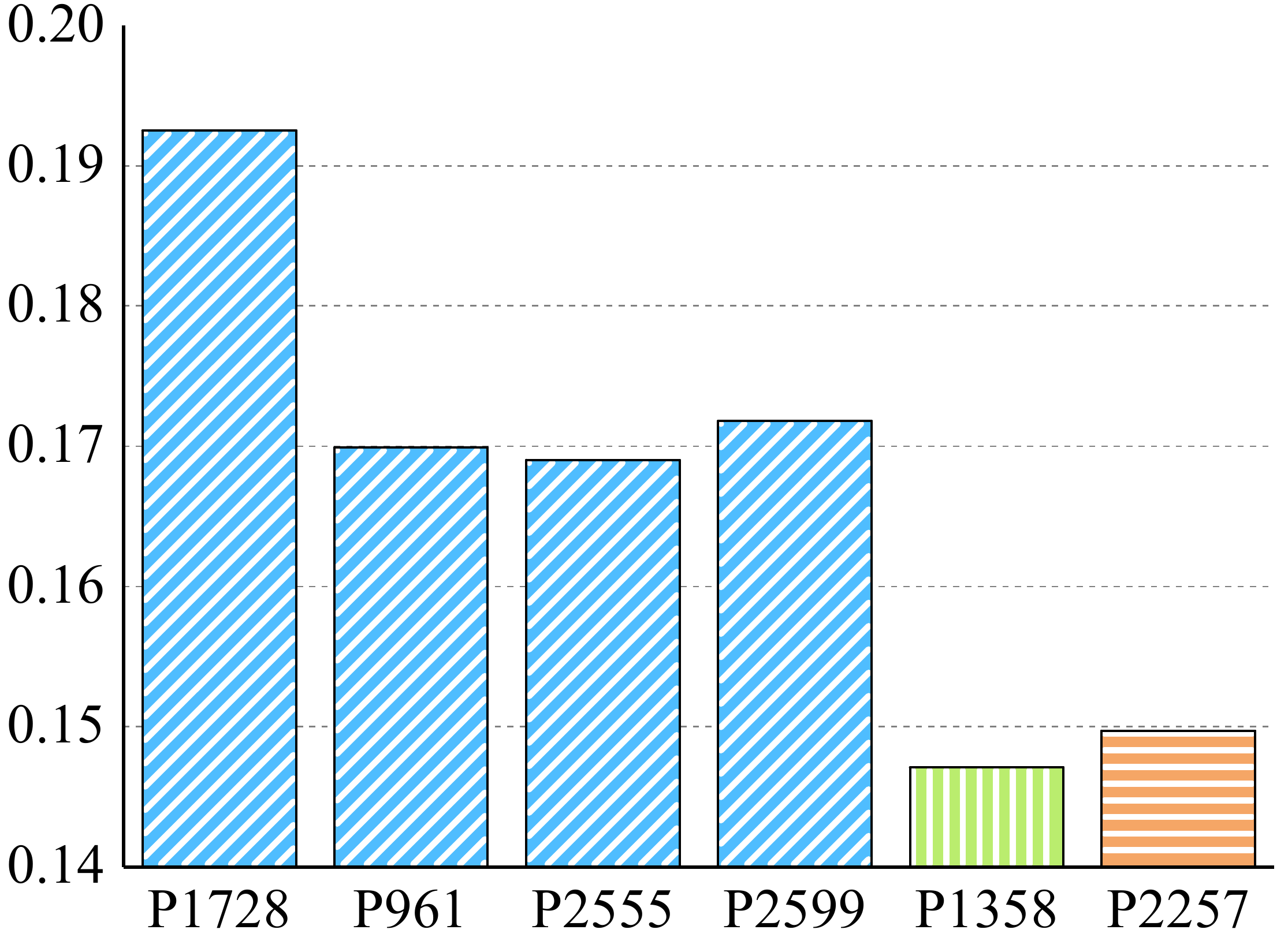}
		\label{fig_attn_2}
	}
	\caption{Neighbors of node P1728 and corresponding attention values. Different colors and patterns indicate different classes.}
	\label{fig_atten}
\end{figure}

\subsubsection{Node Clustering}
Here we conduct the clustering task to evaluate the representation learned from different methods.
For the GNN based methods (i.e., GCN, GAT, and HAT), we can get the feature representations of test nodes from the hidden layer.
Here we utilize K-means to perform node clustering, and the number of clusters is set to the number of labels.
For Poincar\'eEmb and HAT,
we project these representations via $\log_{\mathbf{0}}^c$, and then feed the representations into K-means.
We report the average results of normalized mutual information (NMI)  of 10 runs with random weight initialization.

The results are displayed in Table \ref{table_clustering}.
As we can see, HAT performs better than other baselines in most case, indicating the superior performance of HAT.
Moreover,
for Amazon Photo, 
some graph embedding methods achieve better results than GCN and GAT, 
while HAT still outperforms baselines,
demonstrating the superiority of designing graph neural network in hyperbolic spaces.
Furthermore, the superiority of hyperbolic spaces is further validated in the comparison of LINE(1st) and Poincar\'eEmb.

\subsubsection{Analysis of Attention Mechanism}
We examine the learned attention value of HAT in this section.
Intuitively, more important nodes tend to have larger attention values.
Specifically, we take the paper ``P1728'' in Cora dataset as an illustrative example.
As shown in Fig. \ref{fig_attn_1}, the paper P1728 has 5 neighbors, and the labels of nodes are indicated by colors and patterns.
From Fig. \ref{fig_attn_2}, we can see that the paper P1728 gets the highest attention value,
which means the node itself plays the most essential role in its representation.
P2599, P961 and P2555 get the second, third, fourth highest attention values, respectively.
That is because the three papers belong to the same class with P1728,
and they can make a significant contribution to identifying the class of P1728.
The irrelevant class neighbors, i.e., P1358 and P2257, get the smallest attention values.
Based on the above analysis,
we can see that our proposed attention mechanism can automatically distinguish the difference among neighbors.

\begin{figure}[t]
	\centering
	\subfigure[DeepWalk]{
		\includegraphics[width=0.12\textwidth]{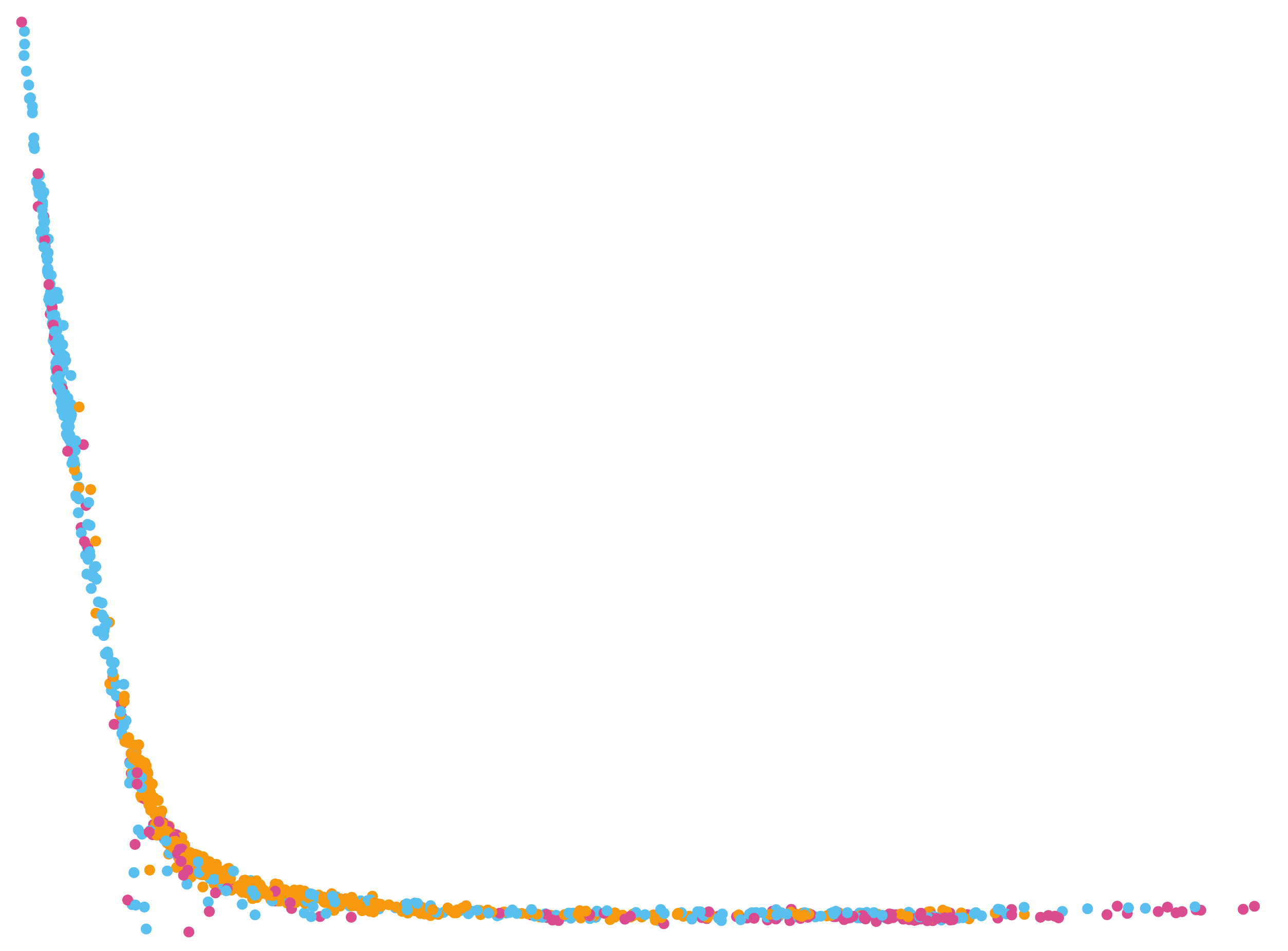}
		\label{fig_visual_deepwalk}
	}
	\subfigure[Node2vec]{
		\includegraphics[width=0.12\textwidth]{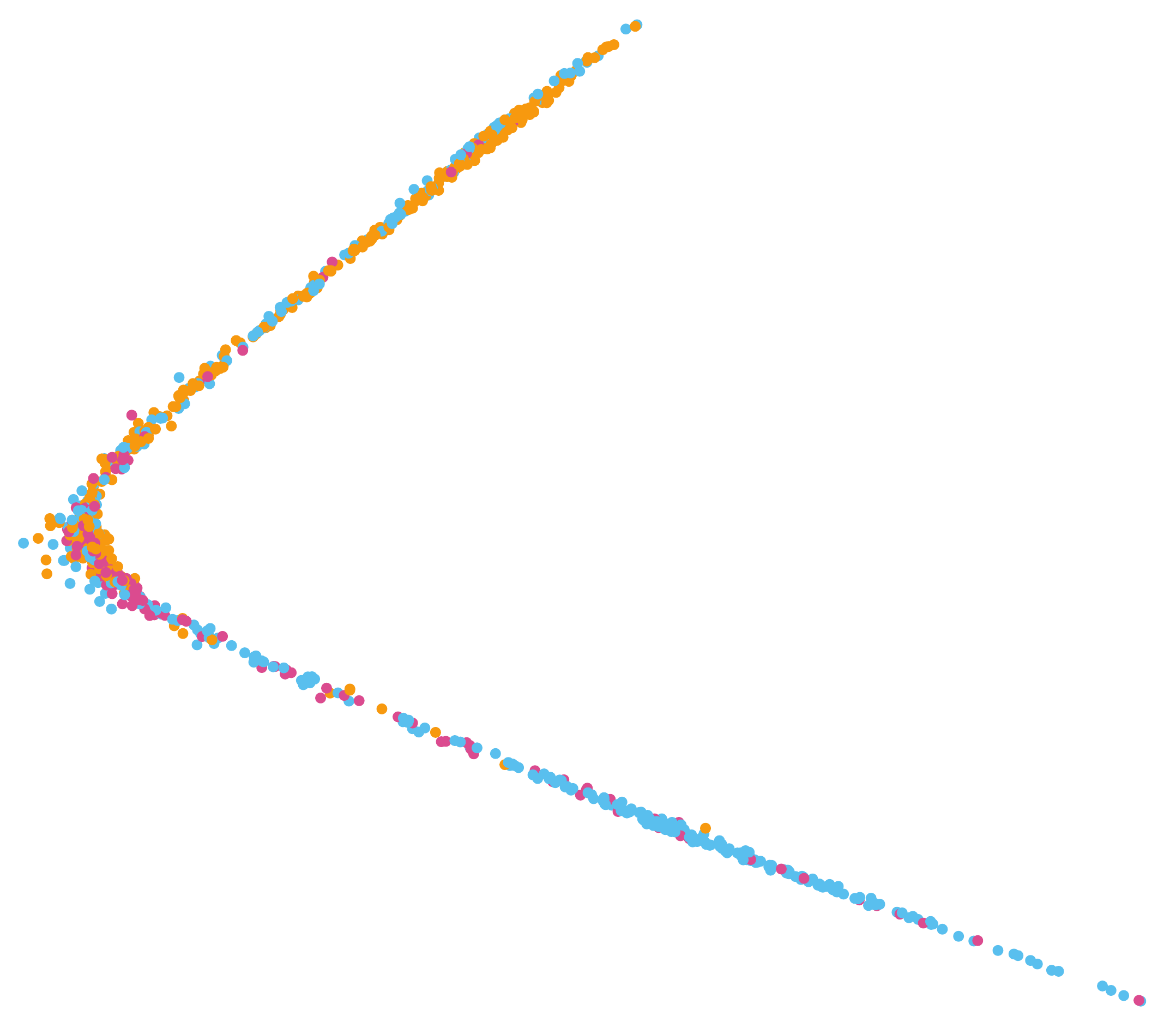}
		\label{fig_visual_node2vec}
	}
	\subfigure[LINE(1st)]{
		\includegraphics[width=0.12\textwidth]{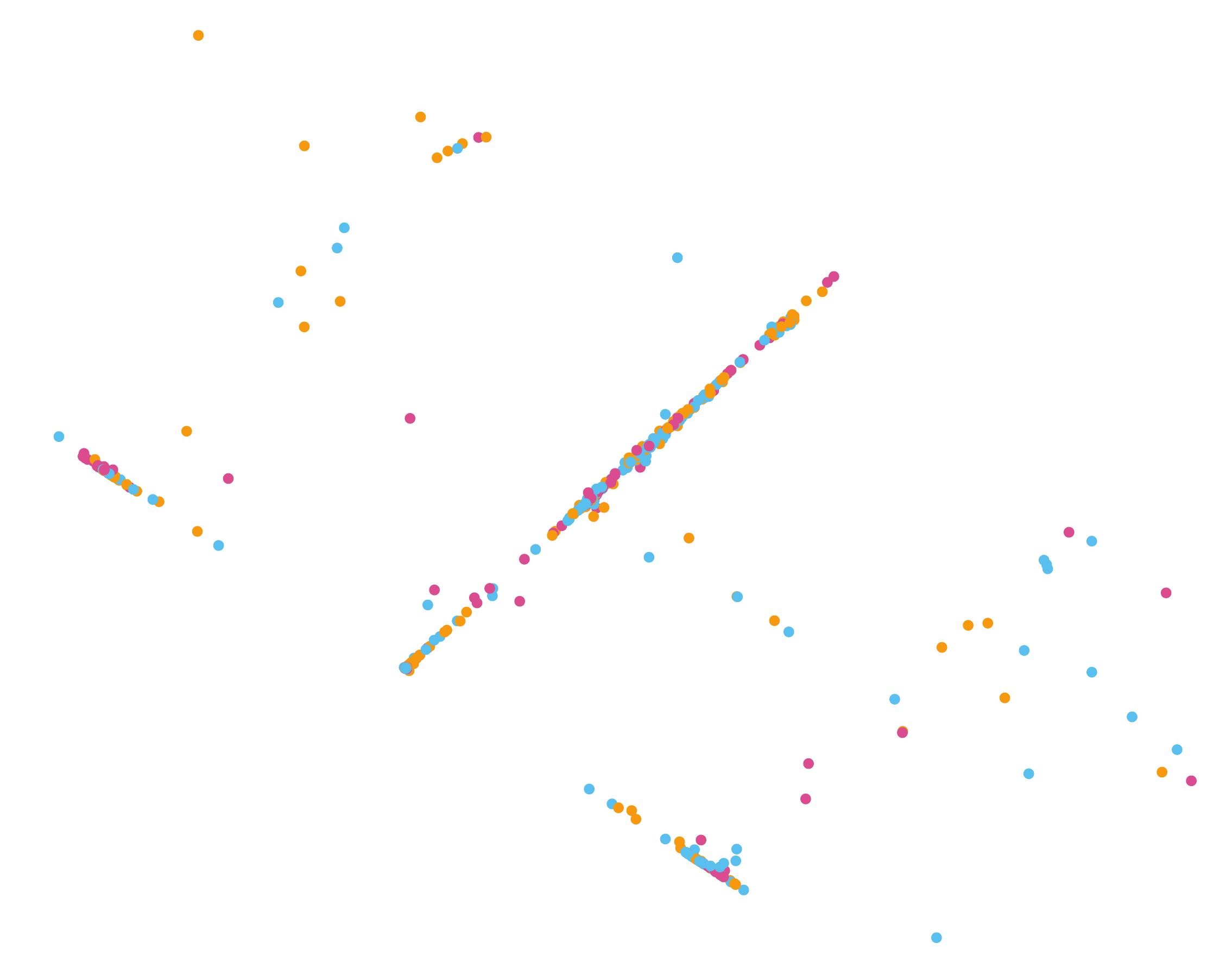}
		\label{fig_visual_line1}
	}
	\subfigure[LINE(2nd)]{
		\includegraphics[width=0.12\textwidth]{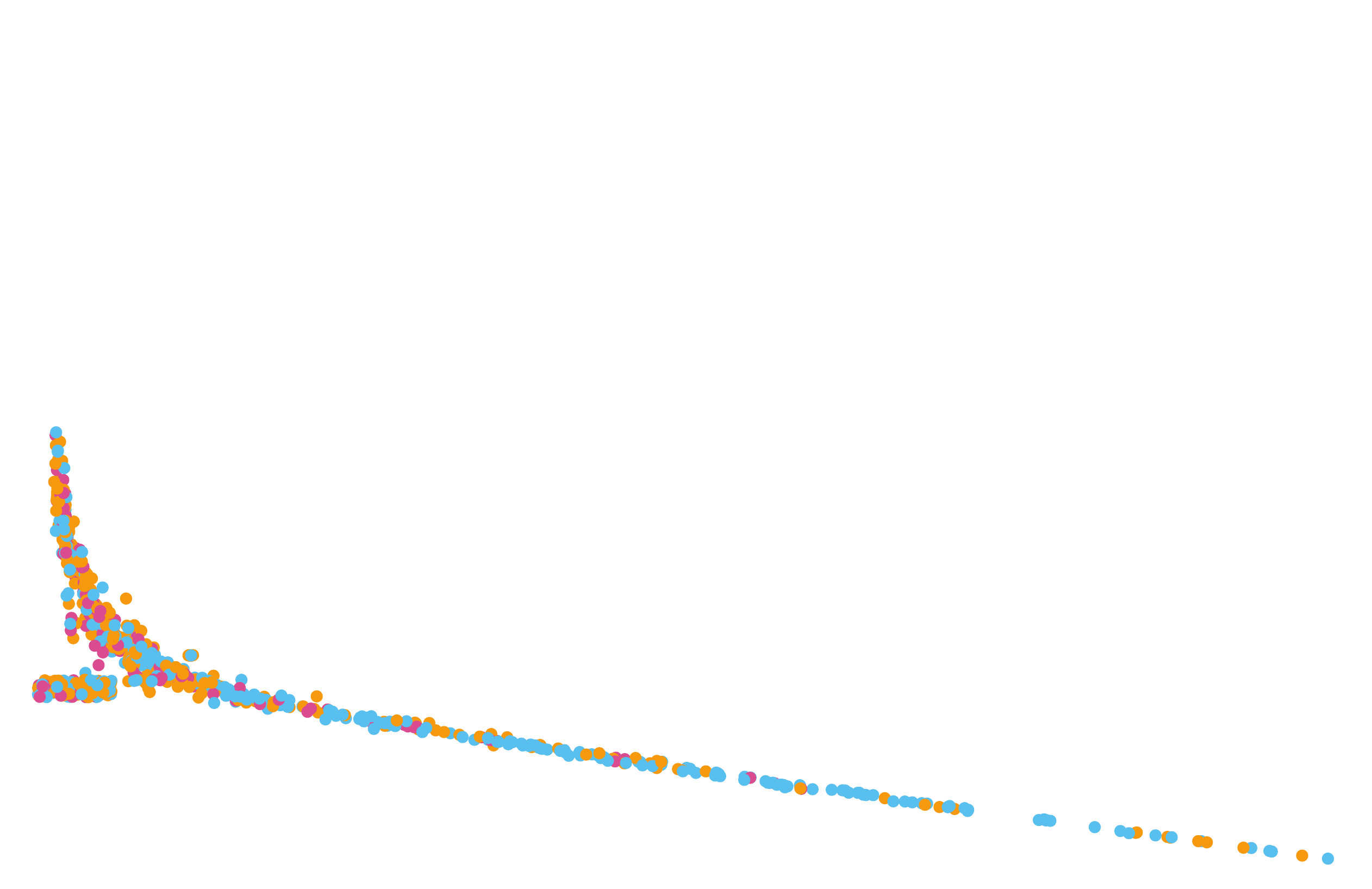}
		\label{fig_visual_line2}
	}
	\subfigure[Poincar\'eEmb]{
		\includegraphics[width=0.12\textwidth]{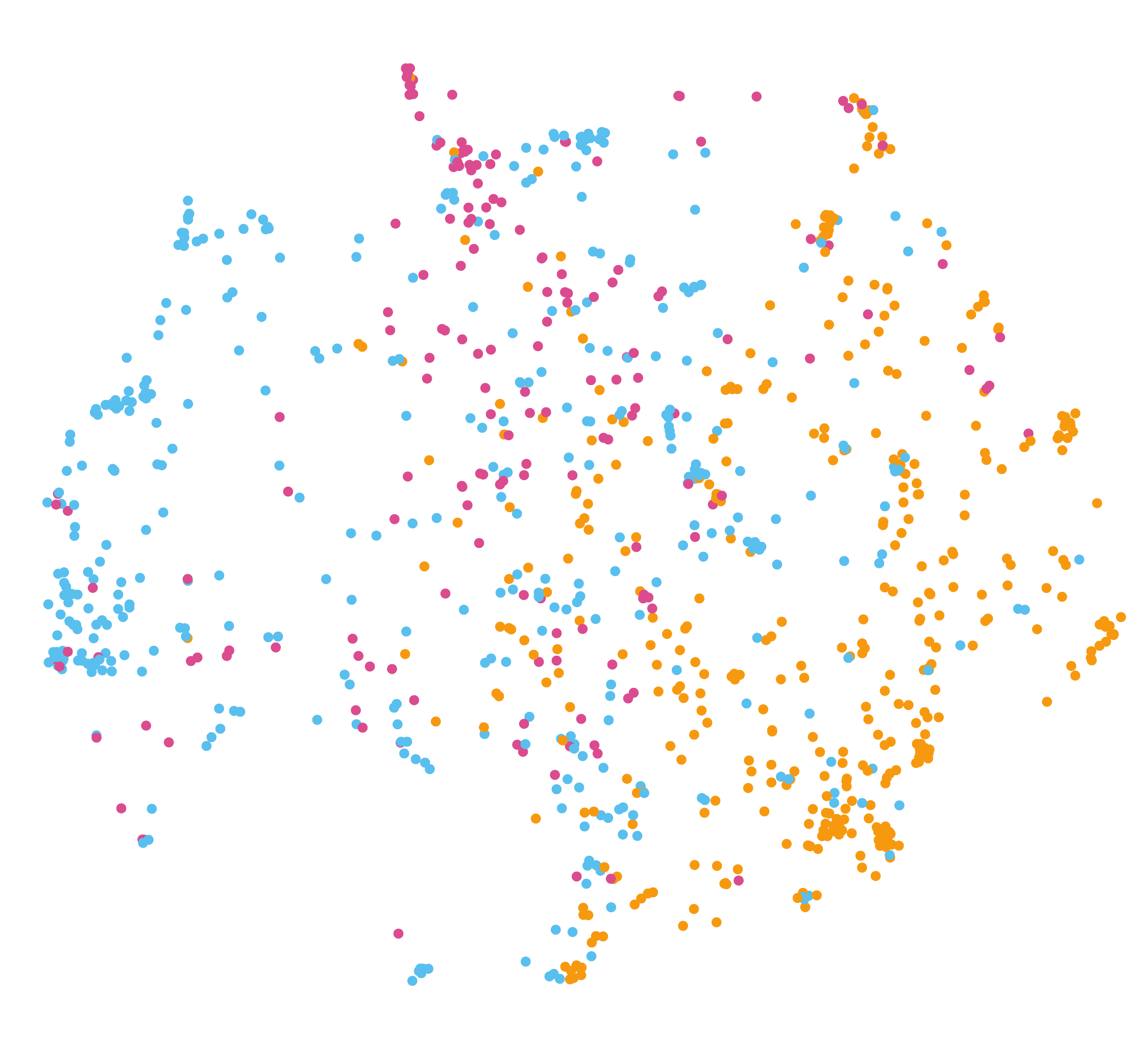}
		\label{fig_visual_poincare}
	}
	\subfigure[GCN]{
		\includegraphics[width=0.12\textwidth]{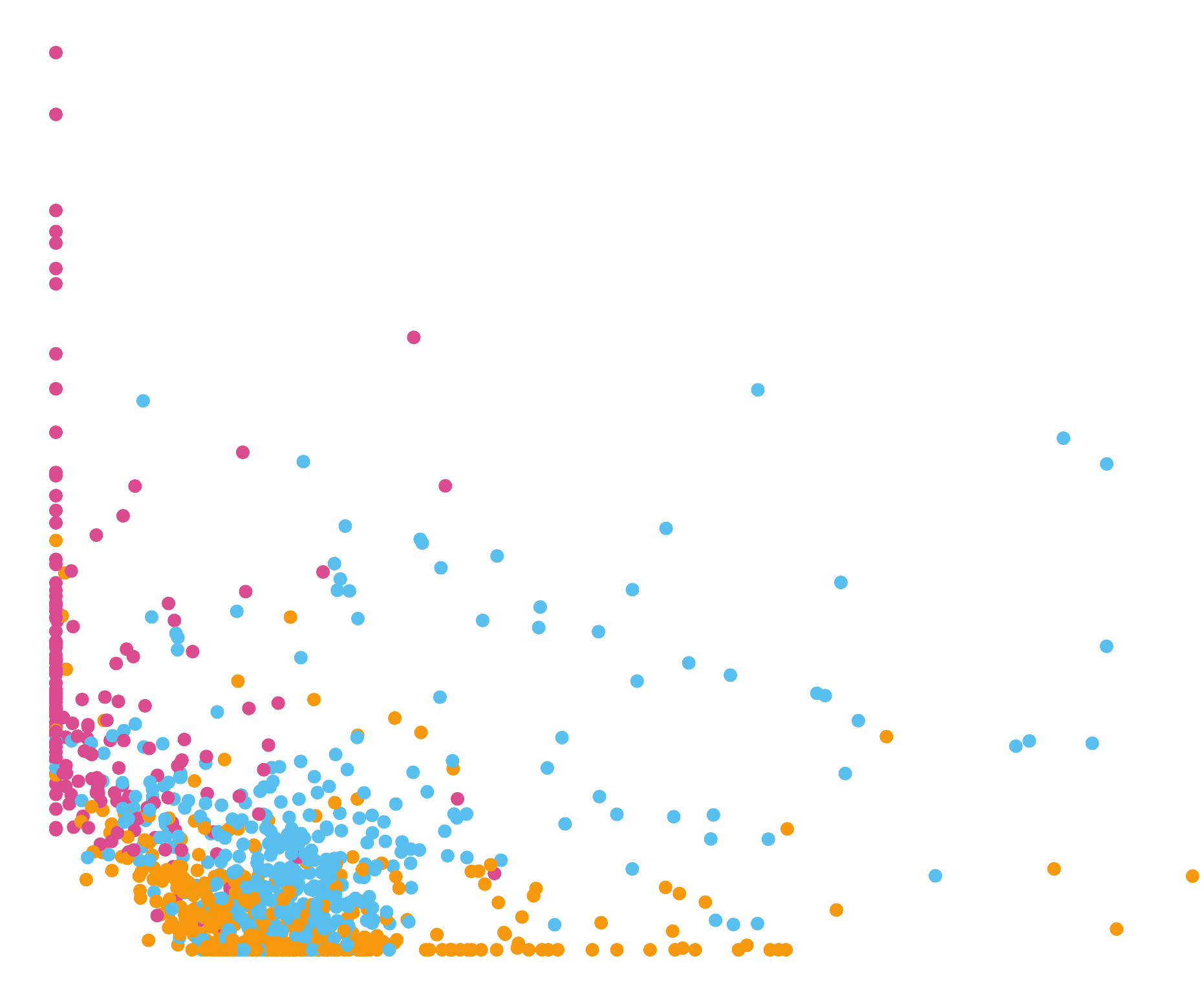}
		\label{fig_visual_gcn}
	}
	\subfigure[GAT]{
		\includegraphics[width=0.12\textwidth]{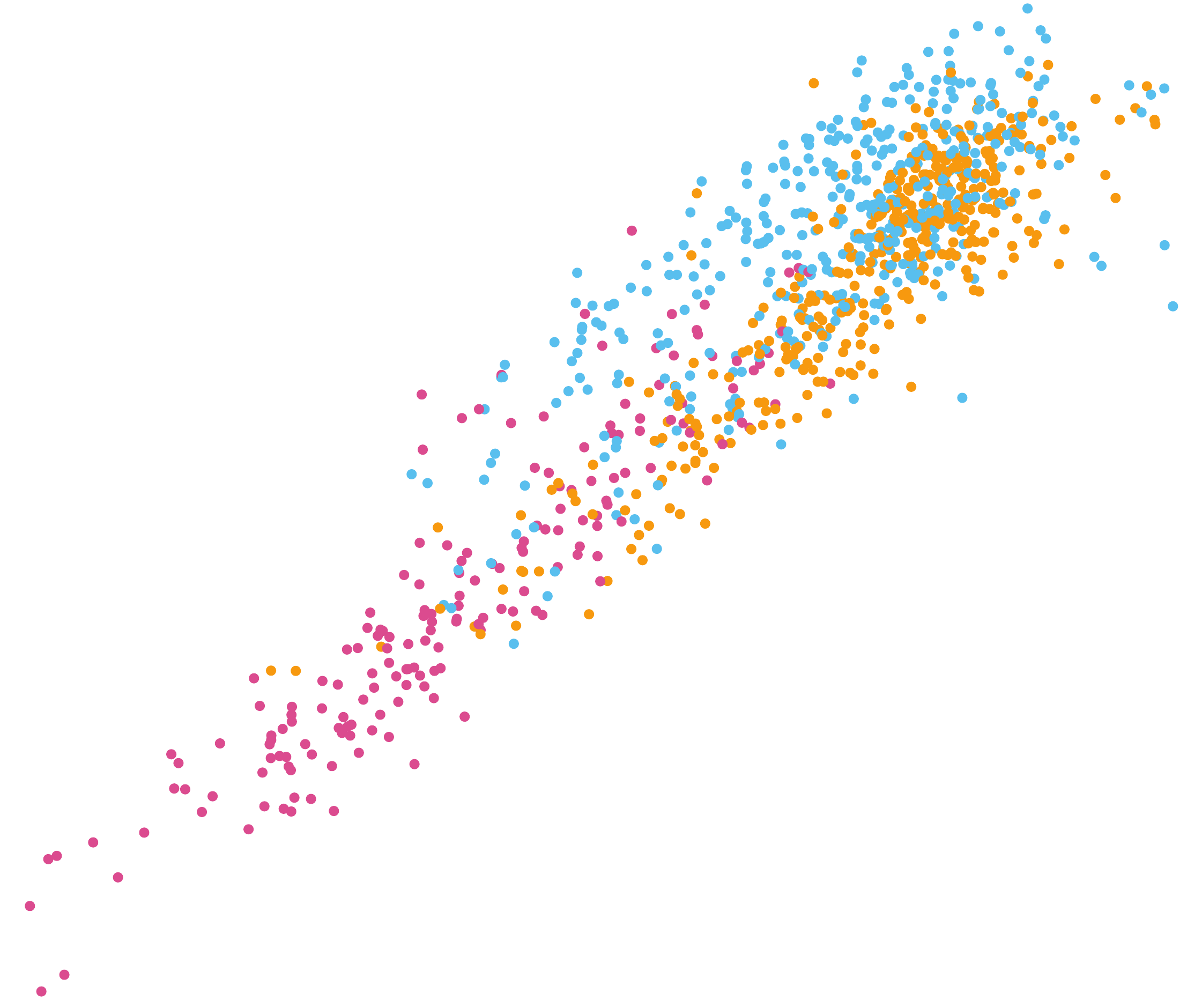}
		\label{fig_visual_gat}
	}
	\subfigure[HAT]{
		\includegraphics[width=0.12\textwidth]{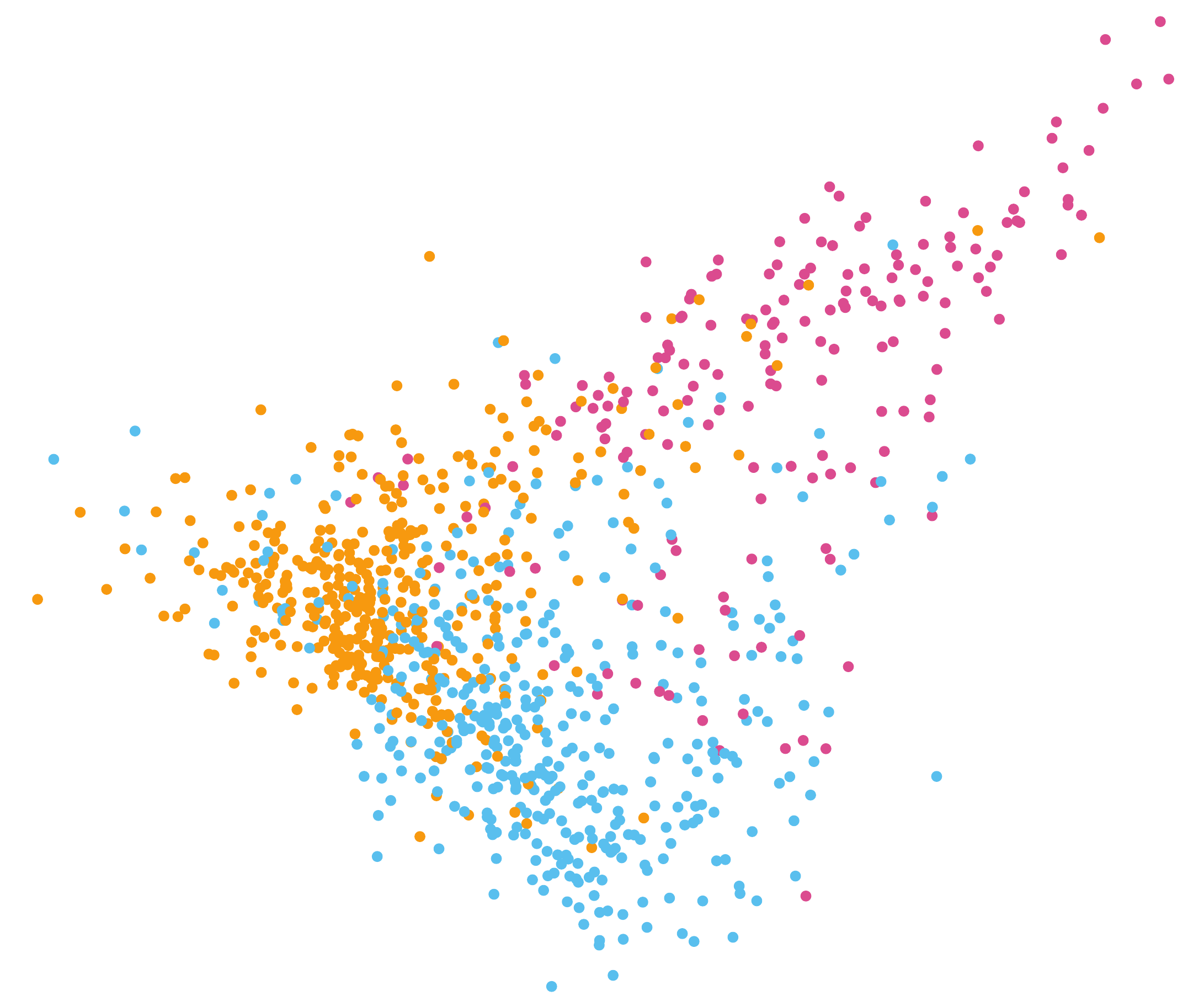}
		\label{fig_visual_hat}
	}
	\caption{Visualization of 2-dimension representations on Pubmed.}
	\label{fig_visual}
\end{figure}

\subsubsection{Graph Visualization}
Graph visualization, aiming to layout a graph on a two-dimensional space,
is another important graph application.
Here, we take Pubmed as a case to visualize the learned representations.
Followed \cite{nickel2017poincare,nickel2018learning,ganea2018hyperbolic}, we directly visualize the learned two-dimensional representations of the nodes.
As shown in Fig. \ref{fig_visual}, each point indicates one paper and its color indicates the label.
We can find that three GNN methods (i.e., HAT, GCN and GAT)
relatively clearly distinguish three classes of nodes.
Compared to GCN and GAT, HAT distinguishes all three categories with a more clear boundary and larger discrimination.


\section{Conclusion}\label{sec_conclusion}

In this paper, we make the first effort toward investigating the graph neural network in hyperbolic spaces and propose a novel hyperbolic graph attention network HAT.
With the framework of gyrovector spaces,
we redesign the graph operations in hyperbolic spaces,
and propose an attention mechanism based on the hyperbolic proximity.
We further devise an acceleration strategy to improve the efficiency of HAT.
The extensive experiments on four datasets demonstrate the superiority of HAT,
compared with the state-of-the-arts.

\section{Acknowledgements}
Hyperbolic representation learning has attracted considerable research attention recently.
We notice that some hyperbolic GNNs were done independently during the same period \cite{chami2019hyperbolic,liu2019hyperbolic,bachmann2019constant},
and I think it is my honor to finish such a relevant work at the same time.
Thank you.

\bibliography{reference}
\bibliographystyle{aaai}

\end{document}